\documentclass[journal]{IEEEtran}
\usepackage{amsmath}
\usepackage{array}
\usepackage{graphicx} 
\usepackage[skip=2pt,font=small]{caption}
\usepackage{xcolor}
\usepackage{colortbl,booktabs}
\usepackage{subfig}
\usepackage{color}
\usepackage{multirow}
\usepackage{multicol}
\usepackage{url}
\graphicspath{{image/}}
\ifCLASSINFOpdf
\else
\fi
\begin{document}
%
\title{Segmentation-Aware and Adaptive Iris Recognition}
\author{Kuo~Wang, Ajay Kumar}
\vspace{-0.5cm}

\maketitle

\begin{abstract}
Iris recognition has emerged as one of the most accurate and convenient biometric for person identification and has been increasingly employed in a wide range of e-security applications. The quality of iris images acquired at-a-distance or under less constrained imaging environments is known to degrade the iris recognition accuracy. The periocular information is inherently embedded in such iris images and can be exploited to assist in the iris recognition under such non-ideal scenarios. Our analysis of such iris templates also indicates significant degradation and reduction in the region of interest, where the iris recognition can benefit from a similarity distance that can consider importance of different binary bits, instead of the direct use of Hamming distance in the literature. Periocular information can be dynamically reinforced, by incorporating the differences in the effective area of available iris regions, for more accurate iris recognition. This paper presents such a periocular-assisted dynamic framework for more accurate less-constrained iris recognition. The effectiveness of this framework is evaluated on three publicly available iris databases using within-dataset and cross-dataset performance evaluation. 
\end{abstract}


%
\IEEEpeerreviewmaketitle

\section{Introduction}\label{I_intro}
%
%
%


\IEEEPARstart{I}{ris} recognition has emerged as one of the most accurate, convenient and low-cost biometric modality to verify the identity of an individual. It is common knowledge \cite{01Daugman2004HowWorks}\cite{02Hollingsworth2009PupilPerformance} that iris patterns are known to be unique among different subjects, even among identical twins, and be easily acquired using low-cost cameras. Therefore iris recognition has been widely incorporated in the national ID programs for the benefit of citizens and effective e-governance. However, the constrained imaging requirements for such widely deployed conventional iris recognition systems, i.e., requirements for the subjects to stop, stand and stare at the iris sensors in the vicinity, poses severe limitations to incorporate iris recognition for the surveillance and forensics. Iris recognition under less-constrained or distantly acquired images has gained increasing importance in recent years. Iris image acquisition module widely use near-infrared (NIR) illumination, typically in the wavelength range of 700-900nm, which can reveal enhanced quality of iris texture under constrained imaging environment. However, with the increase in the standoff distances, the quality of acquired iris images significantly degrades. In such imaging scenarios the periocular information can play an increasingly important role for accurate personal identification. In recent years, periocular recognition has been receiving increasing attention for its promising performance under such less constrained imaging conditions \cite{03Tan2013}\cite{04Park2009PeriocularStudy}. The periocular region usually refers to the region around the eye, which preferably \textit{includes} the eyebrow \cite{05Smereka2015ProbabilisticVerification}. Such periocular near-infrared iris images, in particular, presents highly discriminative features for the person identification. Earlier work \cite{03Tan2013}\cite{04Park2009PeriocularStudy}\cite{06Park2009PeriocularStudy}\cite{07Alonso-Fernandez2018AResearch}\cite{08Zhao2018} in this area has validated that the periocular region is highly discriminative among different persons, and can be considered as an effective alternative or supplement to the face or iris recognition especially when the entire face or clear iris images are not available. This work is motivated to further such advances in the less-constrained iris recognition capabilities and introduces a new framework to more accurately and adaptively match less-constrained iris images.   
\vspace{-0.3cm}
\subsection{Related Work}\label{I.A Related Work}
This section presents a \textit{brief} summary of earlier or related work. We firstly review the related work on iris recognition, followed by the periocular recognition in Section \ref{I.A peri recognition} which also includes promising references on the less-constrained iris recognition. 
\subsubsection{Iris Recognition} \label{I.A iris recognition}
Daugman \cite{01Daugman2004HowWorks} proposed one of the most classic and popular approaches for the automated iris recognition which uses band-pass Gabor filters, on the segmented and normalized iris images, for the feature encoding. These filter responses, including the real-part and imaginary-part, are then binarized to generate \textit{IrisCode} which offers a compact and more robust feature representation. The Hamming distance between two \textit{IrisCodes} is used as the dissimilarity score for verification. Based on \cite{01Daugman2004HowWorks}, 1D log-Gabor filter was incorporated in \cite{09Masek2003} to replace 2D Gabor filter for more efficient iris feature extraction. In 2007, a different approach \cite{10Monro2007DCT-BasedRecognition} using discrete cosine transform (DCT) was explored for analyzing frequency information from fixed-size image blocks and encoded binarized iris features. Miyazawa et al. \cite{11Miyazawa2008AnMatching} propose another spatial-frequency domain approach using 2D discrete Fourier transforms (DFT) which offered promising results. In 2009, Sun and Tan \cite{12ZhenanSun2008} employed the multi-lobe differential filter (MLDF), and referred to as the ordinal filters, which offered an alternative for the Gabor/log-Gabor filters in generating rich iris feature templates.
\par Iris recognition research has also attracted a variety of approaches to enhance segmentation accuracy for the acquired iris images and accurate segmentation is critical in enhancing reliability for the iris recognition. Some of the most widely employed iris segmentation algorithms are based on the integrodifferential operator \cite{01Daugman2004HowWorks} and circular Hough transforms \cite{09Masek2003} which are adapting for detecting iris and pupil circles from the near-infrared eye images. These methods perform well for the high-quality iris images but are quite known to be least reliable for the noisy images acquired under relaxed environments. Tan et.al \cite{13Tan2010EfficientRecognition} proposed an iterative approach to coarsely cluster the iris and non-iris region pixels before applying the integrodifferential operator, and achieved higher reliability in segmenting the iris pixels from noisy iris images. Following the similar coarse-to-fine strategy, a competitive approach is detailed in \cite{03Tan2013} which makes use of the Random Walker algorithm \cite{14Grady2006RandomSegmentation} for coarsely locating the iris region, followed by a couple of gray-level statistics based operators to refine the boundaries. These operators have shown to enable pixel-level precision in the final output or the iris masks. More recent approaches include \cite{15Zhao2015} which utilizes an improved total variation model to address accompanying noise and artifacts in less constrained iris images, and \cite{16Frucci2016WIRE:Recognition} which relies on the color/illumination correction along with the watershed transform for segmenting noisy iris images acquired under visible wavelength. 
\par There has been quite limited work to exploit the potential from deep neural network capabilities for the iris recognition, especially while considering the tremendous popularity of deep learning for various computer vision tasks including for face recognition. An earlier attempt for deep representation of iris appears in \cite{17Menotti2015DeepDetection} in 2015, but such proposal was to detect presentation attacks, a two-class classification problem, instead of the iris recognition. A new approach using \textit{DeepIrisNet} was investigated in \cite{18GangwarDeepIrisNet:RECOGNITION} and used a deep learning-based framework for general iris recognition. This work is essentially a \textit{direct} application of typical convolutional neural networks (CNN) without many optimizations for the iris patterns. Another more recent work in \cite{19He2017DeepNetwork} has attempted to exploit a deep belief net (DBN) for iris recognition. Its core component, however, is the optimal Gabor filter selection, while the DBN is again an application on the \textit{IrisCode} without iris-specific optimization. More recent work in \cite{20Zhao2017TowardsFeatures} proposes a \textit{UniNet} \cite{20Zhao2017TowardsFeatures} employing the deep fully convolutional networks (FCN) \cite{21Long2015} to generate iris binary images and masks for Hamming distance calculation, which explores the substantial connections between iris recognition and deep learning. This work introduces a new loss function that incorporates conventional bit-shifting operations and masks in matching score computations, and achieves state-of-the-art accuracy on several publicly available datasets. Another related and promising work appears in \cite{44proenca2019segmentation} which uses a deep learning architecture to infer misalignment between a pair of iris images that are represented in a segmentation-less polar domain.

\subsubsection{Periocular Recognition} \label{I.A peri recognition}
In recent years, researchers have devoted consistent efforts to investigate new periocular recognition algorithms for the images acquired under less-constrained environments \cite{07Alonso-Fernandez2018AResearch}\cite{22Rattani2017OcularSurvey}. Earlier feasibility study on using the periocular regions for human recognition under varying imaging conditions is undertaken by Park et al. \cite{06Park2009PeriocularStudy} in 2009, and promising results were reported, which provides support for the subsequent research. Bharadwaj et al. \cite{04Park2009PeriocularStudy} further explored the effectiveness of periocular recognition in situations arising from the failure of iris recognition. In this work, part of the later research focuses on cross-spectral periocular matching \cite{23Sharma2014OnRecognition} using the potential from the neural networks. The above explorative works have further motivated the researchers to continuously improve the matching accuracy of periocular images. In 2013, another promising approach appeared in \cite{03Tan2013}, which exploited key-point features and spatial-filter banks, i.e., Dense-SIFT and LMF features, followed by K-means clustering for dictionary learning and representation. However, this approach did not investigate periocular-specific feature representations, and the uses computationally demanding Dense-SIFT features matching. Smereka and Kumar \cite{05Smereka2015ProbabilisticVerification} proposed the Periocular Probabilistic Deformation Model (PPDM) in 2015, which provided sound modeling for the potential deformations that exists among two matched periocular images. Inference of the captured deformation using correlation filter is utilized for matching periocular image pairs. Later in 2016, the same group of researchers improved their basic model by selecting discriminative patch regions for more accountable matching \cite{24Smereka2016SelectingVerification}. These two methods achieved promising performance on multiple datasets. Nevertheless, both of them relied on patch-based matching scheme, and therefore are more susceptible to scale variations or misalignment, that often violate the patch correspondences, which is more likely to happen during the real deployments. Deep learning techniques, especially convolutional neural networks (CNN), have gained immense popularity for computer vision and pattern analysis tasks in recent years.
\par A recent survey on periocular recognition methods \cite{07Alonso-Fernandez2018AResearch} \cite{22Rattani2017OcularSurvey} suggests that few studies have considered the potential from deep learning techniques to boost the periocular matching accuracy. Reference \cite{50hollingsworth2011human} provides insightful observations on periocular features and comparison of machine with human matching performance. In \cite{51bowyer2016handbook}, Bowyer and Burge present a systematic summary on the related ocular recognition systems and algorithms. More recently, Proença and Neves \cite{26Proenca2018Deep-PRWIS:Frameworks} claimed that iris and sclera regions might be less reliable for periocular recognition and proposed Deep-PRWIS. In their work, periocular images are augmented with inconsistent iris and sclera regions for training a deep CNN, so that the network implicitly degrades the iris and sclera features during learning. Promising results were reported from the Deep-PRWIS on two public databases. More promising efforts appear in \cite{08Zhao2018}, which uses a deep learning-based architecture for robust and accurate periocular recognition incorporating the attention model to emphasize the region with higher discriminative information. This algorithm achieves state-of-the-art accuracy on six publicly available databases and can serve as a reasonable baseline for further research in this area.

\begin{table*}[!htbp]
\caption{\normalsize Comparative summary of related and recent work on less-constrained iris recognition.}
\normalsize
\newcommand{\tabincell}[2]{\begin{tabular}{@{}#1@{}}#2\end{tabular}}
\label{table_Summary}
\footnotesize
\centering
\begin{tabular}{m{0cm} m{1cm}| m{1.2cm}<{\centering} |m{1.6cm}<{\centering} |m{2cm}<{\centering} |  m{1.5cm}<{\centering} |m{2cm}<{\centering} | m{1cm}<{\centering} }
\hline
\rule{0pt}{25pt} &  \multirow{2}*{\small\textbf{Ref.}} & \multicolumn{2}{c|}{\small\textbf{Features}} & \small\textbf{Recognition Performance evaluation} & \multicolumn{3}{c}{\small\textbf{\tabincell{c}{Comparative performance \\ for less-constrained iris databases}}}  \\
\cline{3-8}
\rule{0pt}{25pt} & ~ & Iris & Periocular & Recognition & Databases & Recognition rates at FAR=0.0001 & EER \\
\hline
\rule{0pt}{35pt} & \cite{20Zhao2017TowardsFeatures} & Yes & No & No &  \tabincell{l}{(a) \vspace{0.15cm}\\ (b) \vspace{0.15cm} \\  (c)\vspace{0.1cm}}   & \tabincell{l}{38.7\% \vspace{0.15cm}\\ 75.5\% \vspace{0.15cm} \\  75.3\%\vspace{0.1cm}}   & \tabincell{l}{9.73\% \vspace{0.15cm}\\ 3.94\% \vspace{0.15cm} \\  5.54\%\vspace{0.1cm} }  \\
\hline
\rule{0pt}{35pt} & \cite{08Zhao2018} & No & Yes & No & \tabincell{l}{(a) \vspace{0.15cm}\\ (b) \vspace{0.15cm} \\  (c)\vspace{0.1cm}}   & \tabincell{l}{53.9\% \vspace{0.15cm}\\ 64.6\% \vspace{0.15cm} \\  61.6\%\vspace{0.1cm}}   & \tabincell{l}{10.55\% \vspace{0.15cm}\\ 3.93\% \vspace{0.15cm} \\  14.27\%\vspace{0.1cm} }  \\
\hline
\rule{0pt}{35pt} & \cite{03Tan2013} & Yes & Yes & Yes & \tabincell{l}{(a) \vspace{0.15cm}\\ (b) \vspace{0.15cm} \\  (c)\vspace{0.1cm}}   & \tabincell{l}{68.0\% \vspace{0.15cm}\\ 85.4\% \vspace{0.15cm} \\  59.2\%\vspace{0.1cm}}   & \tabincell{l}{13.86\% \vspace{0.15cm}\\ 2.43\% \vspace{0.15cm} \\  9.93\%\vspace{0.1cm} }  \\
\hline
\rule{0pt}{35pt} & \cite{34Zhang2018} & Yes & Yes & No &  \tabincell{l}{(a) \vspace{0.15cm}\\ (b) \vspace{0.15cm} \\  (c)\vspace{0.1cm}}   & \tabincell{l}{44.5\% \vspace{0.15cm}\\ 75.4\% \vspace{0.15cm} \\  21.0\%\vspace{0.1cm}}   & \tabincell{l}{16.74\% \vspace{0.15cm}\\ 7.15\% \vspace{0.15cm} \\  17.99\%\vspace{0.1cm} }  \\
\hline
\rule{0pt}{35pt} & \textbf{Ours} & Yes & Yes & Yes &  \tabincell{l}{(a) \vspace{0.15cm}\\ (b) \vspace{0.15cm} \\  (c)\vspace{0.1cm}}   & \tabincell{l}{83.6\% \vspace{0.15cm}\\ 94.3\% \vspace{0.15cm} \\  86.3\%\vspace{0.1cm}}   & \tabincell{l}{3.87\% \vspace{0.15cm}\\ 0.73\% \vspace{0.15cm} \\  2.29\%\vspace{0.1cm} }  \\
\hline
\rule{0pt}{10pt} & \multicolumn{7}{l}{(a) Q-FIRE \cite{37JohnsonPeterALopez-MeyerPauloSazonovaNadezhdaHuaFSchuckersS2010},     (b) CASIA.v4 Distance \cite{39BiometricsDatabase},     (c) CASIA-Mobile-V1-S3 \cite{34Zhang2018}}
\end{tabular}
\vspace{-0.5cm}
\end{table*}

\subsubsection{Iris and Periocular Feature Fusion} \label{I.A fusion}
The periocular information is simultaneously accessible from the iris images and therefore its use to achieve better iris recognition performance is a feasible strategy. Quite a few prior works have attracted attention to this aspect and several approaches have non-ideal scenarios. In 2010, Woodard et al. \cite{27Woodard2010OnImagery} combined iris and periocular features using score level combination, i.e., weight sum rule, to improve the recognition performance in non-ideal iris imagery. Optimal weights for the two modalities were empirically obtained. Another promising attempt appears in \cite{03Tan2013} which simultaneously recovers the iris feature extracted from log-Gabor filters, periocular features extracted from Dense-SIFT and LMF, to enhance the iris recognition accuracy under relaxed imaging constraints. Raja et al. \cite{28Raja2015Multi-modalPeriocular} propose a framework to combine the information from face, iris and periocular biometric modalities for the user authentication on their smartphones. Various score level combination schemes are explored, including min rule, max rule, product rule, and weighted- score fusion rule, where the weight for each modality is determined according to its contribution to the recognition performance. Besides, some approaches adopt learning-based score-level fusion strategies. Santos et al. \cite{29Santos2015FusingRecognition} present an artificial neural network with two hidden layers to fuse iris and periocular information at the score level for the mobile cross- sensor applications. Verma et al. \cite{30Verma2016At-a-distanceFeatures} utilize the random decision forest (RDF), which is an ensemble learning method, to combine the match scores of iris and periocular biometrics. Noticeable improvement in the performance is shown for at-a-distance person recognition. Ahuja et al. \cite{31Ahuja2016ASpectrum} extract the periocular feature using deep learning and the iris feature using the root SIFT. Then they combine the match scores from these two modalities using the mean rule and linear regression.
There are some other promising attempts in the literature that integrate the information from these two biometric at the decision level and feature level combination. Santos and Hoyle \cite{32talreja2017multibiometric} fuse iris and periocular modality at the decision level to increase the reliability in the unconstrained iris recognition. They train a logistic regression model to predict the weights for each of the classifiers and obtain a final response. Joshi et al. \cite{33Joshi2012PersonBiometrics} investigate iris and periocular biometric performance from their feature level combination. They first concatenate iris and periocular features and then employ the Direct Linear Discriminant Analysis (DLDA) to obtain discriminative and low dimensional feature vectors for the final classification. More recently, Zhang et.al \cite{34Zhang2018} provide a promising framework to combine the iris and periocular features extracted from maxout CNN to enhance the performance for mobile based personal identification.
\vspace{-0.3cm}
\subsection{Our work} \label{I.B our work}
Accuracy of iris recognition under a less-constrained environment is known to significantly degrade, as compared to those from the conventional or standoff iris recognition systems. Such iris images are generally acquired with greater standoff distances, for the surveillance or from the mobile devices with less-cooperative individuals. This research is motivated to address such iris recognition challenges and evaluate iris recognition capabilities under more realistic scenarios. Iris images acquired under less constrained imaging environments often present varying regions of effective iris pixels \cite{01Daugman2004HowWorks}\cite{47cambier2011biometric}. In the context of such less constrained iris images, we revisit the conventional Hamming distance to match binarized iris templates.  Such iris images present significant variations in occlusions which should be carefully considered while simultaneously utilizing available periocular features. Since iris information is inherently embedded in periocular images, the effectiveness of iris matching can benefit from the relative attention or eye area, like for the human visual systems \cite{41mnih2014recurrent}. Table \ref{table_Summary} presents a summary of related work with our work in this paper for the less constrained iris recognition. The key contributions of this paper can be summarized as follows:
\begin{itemize}
    \item This paper introduces a new framework for the periocular assisted iris recognition. Iris images under a less-constrained imaging environment often present varying regions of \textit{effective} iris pixels for the iris matching. Such differences in the effective number of available iris pixels can be used to dynamically reinforce periocular information which is simultaneously available from such iris images. Such dynamic reinforcement should also consider effective regions of discriminative features that receive varying attention during respective periocular matching. Our framework therefore incorporates such discriminative information using a multilayer perceptron network for the less-constrained iris recognition. The experimental results presented in Section \ref{III.B within} for within-database matching using the receiver operating characteristics curve (ROC), on three publicly available databases, indicate outperforming results over state of the art methods. Also, the ROC results presented in Section \ref{III.C cross} show that our algorithm outperforms others in cross-dataset matching. The results from within dataset matching and cross-dataset matching validate the effectiveness and generalization ability of the framework presented in this paper for the less-constrained iris recognition.
    \item The importance of black (0) and white (1) pixels in binarized iris templates may not be the same or similar for iris image templates acquired under less constrained imaging. Therefore this paper presents a new approach to match such templates using a similarity measure, instead of Hamming distance in the literature, which can accommodate the importance of different bits in iris templates. The experimental results presented in this paper on three publicly available iris databases consistently indicate outperforming results and validate the effectiveness of such approach for less-constrained iris recognition.
\end{itemize}
  
Comparative performance from our approach with other competing methods, on three common and public iris images datasets, is also summarized in Table \ref{table_Summary}. The rest of this paper is organized as follows. Section \ref{II_method}, provides details on our unified framework for less constrained iris recognition. This section also includes the architectures for iris and periocular recognition, together with the formulation of the dynamic fusion approach introduced in this work. Our comparative experimental results from within dataset matching and cross-dataset matching using three different public databases are presented in Section \ref{III_Experiment}, The discussion section appears in Section \ref{IV.Discussion} which discusses the theoretical reasons of the effectiveness of our proposed approaches. The key conclusions from this paper are summarized in Section \ref{V.Conclusion}. 

\begin{figure*}[!t]
\centering
\includegraphics[width=6in]{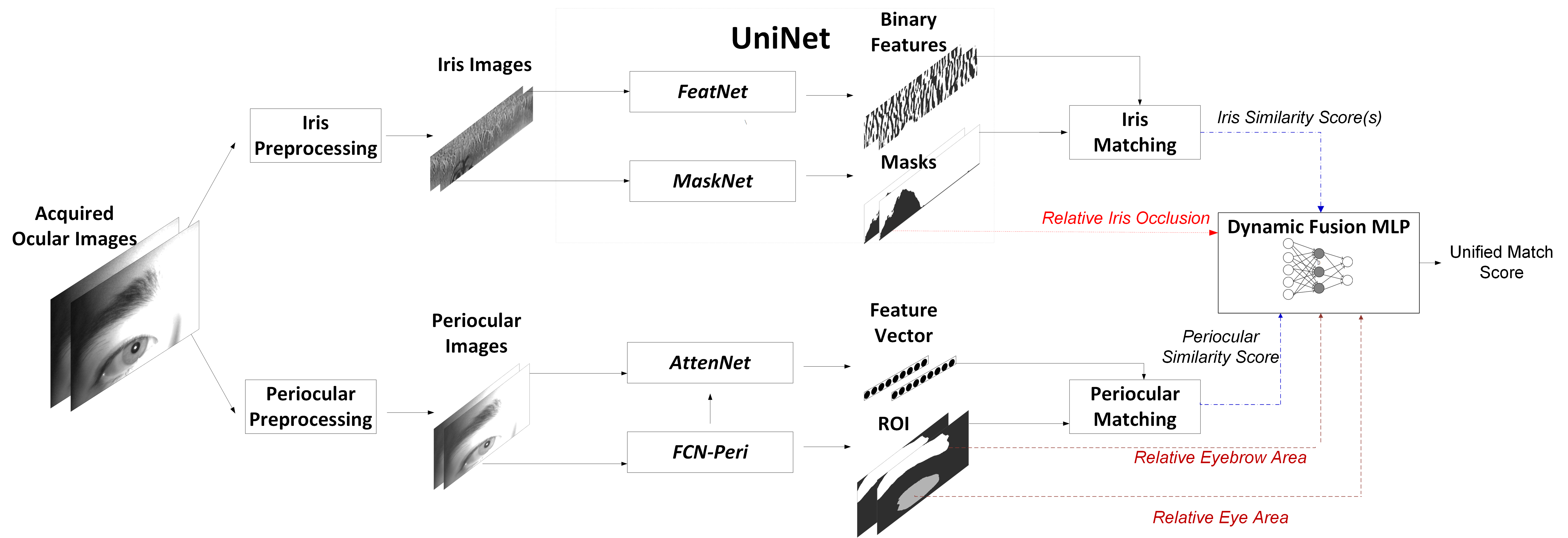}
\caption{The framework for the deep dynamic fusion using iris and periocular information.}
\label{fig_framework}
\end{figure*}

\section{Periocular-Assisted Iris Recognition Framework}\label{II_method}
The framework for periocular-assisted and multi-feature collaboration schemes to achieve dynamic iris recognition is illustrated in Figure \ref{fig_framework}. The detailed explanation of different blocks in this diagram is systematically introduced in the following three sections. This framework adopts the \textit{UniNet} \cite{20Zhao2017TowardsFeatures} to achieve accurate iris matching while the \textit{AttenNet} and \textit{FCN-Peri} \cite{08Zhao2018} are embedded in simultaneously matching the periocular regions in the acquired eye or iris images. The network is trained during two different training or offline phases. We firstly pre-process each of the acquired eye images to independently recover the normalized iris images respective periocular images. The corresponding region of interest images is fed to the respective subnets and trained independently during the first network training phase. During the second training phase, all the parameters in two subnets are frozen and and used to recover recover several cues that indicate the similarity among the iris and periocular templates, including the effective region of iris images among matched template and the corresponding periocular region components among the matched templates. Finally these several cues from the two subnets and employed to train a multilayer perceptron (MLP) network that can enable a binary prediction using the softmax cross-entropy loss. During the performance evaluation or the test phase, a pair of eye images are fed into the trained models and recover the prediction results from the the last softmax layer. These softmax layer results are considered as the consolidated match scores between the input or the unknown eye pair images. Thesse consolidated match scores are used to achieve the binary or the classification decisions for the different applications. Following sections provide further details on different components of the framework.
\vspace{-0.3cm}
\subsection{Iris Template Generation and Comparisons}\label{II.A Iris}
Each of the acquired eye images is first subjected to the localization of region of interest or the iris segmentation and image normalization. These preprocessing steps results in the normalized iris images and were same as employed in earlier work \cite{15Zhao2015}. The dimension of all the segmented and normalized iris images generated from the preprocessing steps, for all the databases employed in our work, is 512$\times$64 pixels. These images are also subjected to the contrast enhancement which saturates 5\% of iris region pixels at high and low intensities. 

The normalized rectangular iris images are subjected to recover respective feature templates and respective masks depicting valid iris pixels or regions. The \textit{UniNet} architecture introduced in \cite{20Zhao2017TowardsFeatures} has shown to offer state-of-art iris matching capabilities and was also adopted in this work. The \textit{UniNet} includes two fully-convolutional sub-networks called \textit{FeatNet} and \textit{MaskNet} as specified in Table \ref{table_U2niNet}. The \textit{MaskNet} generates binary mask distinguishing the valid and invalid or less reliable regions in the iris templates that often degrade the iris matching accuracy. The network uses triplet architecture for the training and we generate triplets in a ratio of 1:3 between the genuine match pairs and the imposter match pairs for the respective training sets. 
The \textit{MaskNet} is pre-trained using from ND-IRIRS-0405 Iris Image Dataset \cite{35Phillips2010} and all the parameters are frozen in this work. The \textit{FeatNet}, pretrained with ND-IRIRS-0405 Iris Image Dataset and publicly made available from \cite{20Zhao2017TowardsFeatures}, is finetuned using the triplet pairs generated from the respective training sets. The \textit{FeatNet} is essentially a fully convolutional neural network and aims to learn the same size but more robust pseudo-binary representation of the input iris images. The loss function introduced in the \textit{FeatNet} training is the extended triplet loss which aims to enlarge the margin of the pseudo-Hamming distance between the intra-class and inter-class matching. The extended triplet loss $L_{ETL}$ can be defined as follows.

\begin{table}[!t]
\caption{The specification of incorporated \textit{\textit{UniNet}}.}
\label{table_U2niNet}
\centering
\begin{tabular}{| m{1.5cm}<{\centering} |m{2cm}<{\centering} |m{1.4cm}<{\centering} | m{1.6cm}<{\centering} | }
\hline
\textbf{Layer Name} & \textbf{Layer Type} & \textbf{Kernel Size} & \textbf{Output Channel} \\
\hline
\multicolumn{4}{|c|}{\textbf{\textit{FeatNet}}}\\
\hline
\vspace{2pt}
Conv1 & Convolution & 3$\times$7 & 16  \\
\hline
\vspace{2pt}
Tanh1 & TanH Activation & - & 16  \\
\hline
\vspace{2pt}
Pool1 & Average Pooling & 2$\times$2 & 16  \\
\hline
\vspace{2pt}
Conv2 & Convolution & 3$\times$5 & 32  \\
\hline
\vspace{2pt}
Tanh2 & TanH Activation & - & 32  \\
\hline
\vspace{2pt}
Pool2 & Average Pooling & 2$\times$2 & 32  \\
\hline
\vspace{2pt}
Res1 & Deconvolution & 4$\times$4 & 32  \\
\hline
\vspace{2pt}
Conv3 & Convolution & 3$\times$3 & 64  \\
\hline
\vspace{2pt}
Tanh3 & TanH Activation & - & 64  \\
\hline
\vspace{2pt}
Pool3 & Average Pooling & 2$\times$2 & 64  \\
\hline
\vspace{2pt}
Res2 & Deconvolution & 8$\times$8 & 64  \\
\hline
\vspace{2pt}
Concat & Concatenation & - & 112  \\
\hline
\vspace{2pt}
Conv4 & Convolution & 3$\times$3 & 1  \\
\hline
\multicolumn{4}{|c|}{\textbf{\textit{MaskNet}}} \\
\hline
m\_Conv1 & Convolution & 3$\times$3 & 16  \\
\hline
\vspace{2pt}
m\_ReLU1 & ReLU Activation & - & 16  \\
\hline
\vspace{2pt}
m\_Pool1 & Max Pooling & 2$\times$2 & 16  \\
\hline
\vspace{2pt}
m\_Conv2 & Convolution & 3$\times$5 & 32  \\
\hline
\vspace{2pt}
m\_ReLU2 & ReLU Activation & - & 32  \\
\hline
\vspace{2pt}
m\_Pool2 & Max Pooling & 2$\times$2 & 32  \\
\hline
\vspace{2pt}
m\_Score2 & Convolution & 1$\times$1 & 2  \\
\hline
\vspace{2pt}
m\_Conv3 & Convolution & 3$\times$5 & 64  \\
\hline
\vspace{2pt}
m\_ReLU3 & ReLU Activation & - & 64  \\
\hline
\vspace{2pt}
m\_Pool3 & Max Pooling & 2$\times$2 & 64  \\
\hline
\vspace{2pt}
m\_Score3 & Convolution & 1$\times$1 & 2  \\
\hline
\vspace{2pt}
m\_Conv4 & Convolution & 3$\times$5 & 128  \\
\hline
\vspace{2pt}
m\_ReLU4 & ReLU Activation & - & 128  \\
\hline
\vspace{2pt}
m\_Pool4 & Max Pooling & 2$\times$2 & 128  \\
\hline
\vspace{2pt}
m\_Score4 & Convolution & 1$\times$1 & 2  \\
\hline
\vspace{2pt}
m\_Upscore4 & Deconvolution & 8$\times$8  & 2  \\
\hline
\vspace{2pt}
m\_Score34 & Elementwise Sum & - & 2  \\
\hline
\vspace{2pt}
m\_Upscore34 & Deconvolution & 4$\times$4  & 2  \\
\hline
\vspace{2pt}
m\_Score234 & Elementwise Sum & - & 2  \\
\hline
\vspace{2pt}
m\_Fuse & Deconvolutiion & 4$\times$4 & 2  \\
\hline
\end{tabular}
\vspace{-0.5cm}
\end{table}

\begin{equation}
\label{eq1}
\small
L_{ETL}=\frac{1}{N}\sum_{i=1}^{N}{max(||M_P-M_A||^2-||M_N-M_A||^2+m, 0)}
\end{equation}
where $N$ is the batch size, $M_A$,$M_P$,$M_N$ are the corresponding masked feature maps generated by the \textit{FeatNet}, and m is the hyperparameter controlling the margin between anchor-positive and anchor-negative distances. 
\vspace{-0.3cm}
\subsection{Comparisons using Similarity Score}\label{II.B WS}
Hamming distance is widely employed to compute the dissimilarities between two binary feature templates in a range of biometric identification problems, such as for the iris or the palmprint recognition. It assumes that the information content from all the template values in the coding space is equally important to distinguish the user identity. However the choices of feature extraction and binarization methods, along with the nature of input images, can effectively determine the importance of white (ones) area and black (zeros) area in the encoded images. Therefore, a more flexible distance measure that can consider such asymmetric importance is proposed to be incorporated for matching less-constrained iris images. Such measure is also referred to as the weighted similarity score (WS) with azzoo similarity measure \cite{36Cheng2018} and was also incorporated for matching iris templates. 
\par The effectiveness of white pixel matching and the black pixel matching in feature templates can also be experimentally evaluated. Let us assume that the number of white pixels and black pixels from one feature template A can be respectively represented as $P_W (A)$ and $P_B (A)$. While comparing two template A and template B, we can perform only white pixels matching $M_W(A,B)$ and only black pixels matching $M_B(A,B)$, and can compute the white pixel matching rates $R_w(A,B)$ and black pixel matching rate $R_B(A,B)$ as shown in the following two equations.
\begin{equation}
\label{eq2}
\small
R_{W}(A,B)=\frac{2\times M_{W}(A,B)}{P_{W}(A)+P_{W}(B)}
\end{equation}
\begin{equation}
\label{eq3}
\small
R_{B}(A,B)=\frac{2\times M_{B}(A,B)}{P_{B}(A)+P_{B}(B)}
\end{equation}
The difference in the contributions from different pixels matching,i.e. average $R_W$ and $R_B$ from the genuine and imposter pairs, can also be empirically observed from the experiments using templates generated from the databases. We select 1,000 genuine matching and 2,000 imposter matching from the test on CASIA-Mobile-V1-S3 dataset for empirical evaluation. It was observed that the average $R_W$ are 0.5733 and the average $R_B$ is 0.6138 for the genuine matches, while $R_W$ are 0.4159 and the average $R_B$ is 0.4563 for the imposter matches
\par In order to accommodate differences in the discriminative information from the white pixel pairs and from the black pixel pairs, we use different weight and generate weighted similarity measure as follows:
\begin{equation}
WS(I_{i,j}^{1},I_{i,j}^{2}) =\left\{
\begin{array}{rcl}
2-\alpha & & {if\quad I_{i,j}^{1} = I_{i,j}^{2} = 1} \\
\alpha & & {if\quad I_{i,j}^{1} = I_{i,j}^{2} = 0}\\
0 & & {if\quad I_{i,j}^{1} \neq I_{i,j}^{2}}
\end{array} \right.
\end{equation}
where $I_{i,j}^{1}$,$I_{i,j}^{2}$ are pixels in row i and column j in two matched two iris templates, and $\alpha$ is hyperparameter controlling the significance of coding pairs. In all our experiments, $\alpha$ is empirically set as 0.3. Assuming the image size of iris images are $H\times W$, we generate the match score using the weighted similarity as follows:

\begin{equation}
\label{eq5}
\small
 S_{WS}=\frac{1}{H\times W}\sum_{i=1}^{H}\sum_{j=1}^{W}WS(I_{i,j}^{1},I_{i,j}^{2}) 	
\end{equation}
It can be observed that when the $\alpha$ is unity, the value of $S_{ws}$ is essentially the difference between unity and the normalized Hamming distance. Therefore weighted similarity can be considered as a more flexible alternative for the templates matching.

\subsection{Periocular Template Generation and Comparisons}\label{II.C Periocular}
\begin{table}[!t]
\caption{Details on the architecture for the \textit{AttenNet} and \textit{FCN-Peri}.}
\label{table_AttenNet}
\centering
\begin{tabular}{| m{3cm}<{\centering} |m{2cm}<{\centering} |m{1cm}<{\centering} | m{1.2cm}<{\centering} | }
\hline
\textbf{Layer Name} & \textbf{Layer Type} & \textbf{Kernel Size} & \textbf{Output Channel} \\
\hline
\multicolumn{4}{|c|}{\textbf{\textit{AttenNet}}}\\
\hline
\vspace{2pt}
Conv1\_1 & Convolution & 5$\times$5 & 32  \\
\hline
\vspace{2pt}
ReLU1\_1 & ReLU Activation & - & 32  \\
\hline
\vspace{2pt}
Conv1\_2 & Convolution & 5$\times$5 & 32  \\
\hline
\vspace{2pt}
ReLU1\_2 & ReLU Activation & - & 32  \\
\hline
\vspace{2pt}
Pool1 & Max Pooling & 2$\times$2 & 32  \\
\hline
\vspace{2pt}
Slice\_roi & Slice & - & 32  \\
\hline
\vspace{2pt}
Conv2\_1, A\_Conv2\_1 & Convolution & 3$\times$3 & 32  \\
\hline
\vspace{2pt}
ReLU2\_1, A\_ReLU2\_1 & ReLU Activation & - & 32  \\
\hline
\vspace{2pt}
Conv2\_2, A\_Conv2\_2 & Convolution & 3$\times$3 & 32  \\
\hline
\vspace{2pt}
ReLU2\_2, A\_ReLU2\_2 & ReLU Activation & - & 32  \\
\hline
\vspace{2pt}
Pool2, A\_Pool2 & Max Pooling & 2$\times$2 & 32  \\
\hline
\vspace{2pt}
Att2 & Attention & - & -  \\
\hline
\vspace{2pt}
Conv3\_1, A\_Conv3\_1 & Convolution & 3$\times$3 & 64  \\
\hline
\vspace{2pt}
ReLU3\_1, A\_ReLU3\_1 & ReLU Activation & - & 64  \\
\hline
\vspace{2pt}
Conv3\_2, A\_Conv3\_2 & Convolution & 3$\times$3 & 64  \\
\hline
\vspace{2pt}
ReLU3\_2, A\_ReLU3\_2 & ReLU Activation & - & 64  \\
\hline
\vspace{2pt}
Pool3, A\_Pool3 & Max Pooling & 2$\times$2 & 64  \\
\hline
\vspace{2pt}
Conv4\_1, A\_Conv4\_1 & Convolution & 3$\times$3 & 64  \\
\hline
\vspace{2pt}
ReLU4\_1, A\_ReLU4\_1 & ReLU Activation & - & 64  \\
\hline
\vspace{2pt}
Conv4\_2, A\_Conv4\_2 & Convolution & 3$\times$3 & 64  \\
\hline
\vspace{2pt}
ReLU4\_2, A\_ReLU4\_2 & ReLU Activation & - & 64  \\
\hline
\vspace{2pt}
Pool4, A\_Pool4 & Max Pooling & 2$\times$2 & 64  \\
\hline
\vspace{2pt}
Att4 & Attention & - & -  \\
\hline
\vspace{2pt}
Feat, A\_Feat & Fully Connecte& - & 64  \\
\hline
\multicolumn{4}{|c|}{\textbf{\textit{FCN-Peri}}} \\
\hline
\vspace{2pt}
Conv1 & Convolution & 5$\times$5 & 16  \\
\hline
\vspace{2pt}
ReLU1 & ReLU Activation & - & 16  \\
\hline
\vspace{2pt}
Pool1 & Max Pooling & 2$\times$2 & 16  \\
\hline
\vspace{2pt}
Conv2 & Convolution & 3$\times$3 & 32  \\
\hline
\vspace{2pt}
ReLU2 & ReLU Activation & - & 32  \\
\hline
\vspace{2pt}
Conv2\_s & Convolution & 1$\times$1 & 3  \\
\hline
\vspace{2pt}
Pool2 & Max Pooling & 2$\times$2 & 32  \\
\hline
\vspace{2pt}
Conv3 & Convolution & 3$\times$3 & 64  \\
\hline
\vspace{2pt}
ReLU3 & ReLU Activation & - & 64  \\
\hline
\vspace{2pt}
Conv3\_s & Convolution & 1$\times$1 & 3  \\
\hline
\vspace{2pt}
Pool3 & Max Pooling & 2$\times$2 & 64  \\
\hline
\vspace{2pt}
Conv4 & Convolution & 3$\times$3 & 128  \\
\hline
\vspace{2pt}
ReLU4 & ReLU Activation & - & 128  \\
\hline
\vspace{2pt}
Conv4\_s & Convolution & 1$\times$1 & 3  \\
\hline
\vspace{2pt}
Upscore4 & Deconvolution & 8$\times$8 & 3  \\
\hline
\vspace{2pt}
Score34 & Elementwise Sum & - & 3  \\
\hline
\vspace{2pt}
Upscore34 & Deconvolution & 4$\times$4 & 3  \\
\hline
\vspace{2pt}
Score234 & Elementwise Sum & - & 3  \\
\hline
\vspace{2pt}
Fuse & Deconvolution & 4$\times$4 & 3  \\
\hline
\end{tabular}
\end{table}
The periocular preprocessing is more simplified and incorporates image normalization with a bilinear filter. The dimensions of all normalized periocular images are empirically fixed as 300$\times$240. Earlier research has shown that periocular recognition with attention models can offer state-of-art performances \cite{08Zhao2018} and was also employed for generating periocular template images for the matching. Therefore the periocular recognition model also includes two components, \textit{FCN-Peri} and \textit{AttenNet}. The architecture for these networks are detailed in Table \ref{table_AttenNet}. The \textit{FCN-Peri} is a fully convolutional network which aims to detect the eye region and eyebrow region in the presented periocular images. We use the \textit{FCN-Peri} for the near-infrared (NIR) images, as publicly made available in \cite{08Zhao2018}, and do not perform any further tuning. With such automatically detected eye and eyebrow region, the \textit{AttenNet} provides pixel locations to these specific particular regions so that specific attention is incorporated to these locations in generating more discriminant periocular features. The output of \textit{AttenNet} is a feature vector with 512 elements. We compute the distance-driven sigmoid cross-entropy (DSC) loss between the siamese pairs, which are generated from the corresponding training set during the training phase. The ratio of genuine pairs and imposter pairs is set empirically set as 1:2 for all our experiments. The DSC loss $L_{DSC}$ \cite{08Zhao2018} can be defined as follows.  
\begin{equation}
\label{eq6}
\small
L_{DSC}=-\frac{1}{N}\sum_{i=1}^{N}(t\log{(\frac{1}{1-e^{-s}})}+(1-t)\log(\frac{e^{-s}}{1-e^{-s}}))
\end{equation}
where $N$ is the batch size, $t$ is the ground truth label for every genuine and imposter pair, and $s$ is a transformed Euclidean distance. 
\subsection{Segmentation-Aware Dynamic Fusion }
\begin{table}[!t]
\caption{The specification of incorporated MLP.}
\label{table_MLP}
\centering
\begin{tabular}{| m{1.5cm}<{\centering} |m{2cm}<{\centering} |m{1.4cm}<{\centering} | m{1.6cm}<{\centering} | }
\hline
\textbf{Layer Name} & \textbf{Layer Type} & \textbf{Input Channel} & \textbf{Output Channel} \\
\hline

\vspace{2pt}
FC1 & Fully Connected & 8 & 32  \\
\hline
\vspace{2pt}
Tanh1 & TanH Activation & 32 & 32  \\
\hline
\vspace{2pt}
FC2 & Fully Connected & 32 & 16  \\
\hline
\vspace{2pt}
Tanh2 & TanH Activation & 16 & 16  \\
\hline
\vspace{2pt}
FC3 & Fully Connected & 16 & 8  \\
\hline
\vspace{2pt}
Tanh3 & TanH Activation & 8 & 8  \\
\hline
\vspace{2pt}
FC4 & Fully Connected & 8 & 2  \\
\hline
\end{tabular}
\vspace{-0.5cm}
\end{table}
Any effective dynamic mechanism to simultaneously utilize the iris and periocular information should carefully consider multiple cues, not just from the individual feature similarity but also from the segmentation steps which can provide (dynamic) importance for the individual similarity scores. Iris images under less-constrained imaging often present varying number of \textit{effective} iris pixels, that are incorporated to generate respective iris match scores. The differences in the effective number of available iris pixels, among two matched iris images, can be used to dynamically reinforce periocular information for more reliable match score. Such dynamic reinforcement should also consider effective regions of discriminative features, which are receiving \textit{varying} attention during respective periocular matching. Therefore we incorporate multilayer perceptron network to dynamically consolidate such multiple pieces of discriminative information and generate more reliable consolidated match score between two unknown or input images.
\par As illustrated in Figure \ref{fig_framework}, the \textit{UniNet} generates pseudo-binary feature maps, along with the respective masks, while the \textit{AttenNet} generates the feature vectors to compute Euclidean distance among respective ROI maps. Therefore we can simultaneously generate iris match scores and periocular match score using the Euclidean distance. Another important input for MLP, which effectively represents the importance or the quality of respective iris match scores, is the mask rate. This mask rate is the ratio between the valid pixels and all iris pixels among two matched iris image templates. Similarily the effectiveness of periocular feature template match scores is represented using the eye and eyebrow ratio sum and the difference, i.e., sum (also difference) of \textit{eye} areas among matched periocular images and sum (also difference) of \textit{eyebrow} areas among matched periocular images. It should be noted that these eye and eyebrow areas are automatically predicted or available from \textit{AttenNet} as shown in Figure \ref{fig_framework}. The MLP network therefore receives an eight-element feature vector and is trained offline using respective genuine and impostor pairs from the training dataset. The architecture of incorporated MLP is shown in Table \ref{table_MLP}. The network training attempts for a binary classification using softmax cross-entropy loss, with respective genuine and impostor class labels. The trained network is used to generate consolidated match scores from the softmax value in the last layer output which ranges between 0 and 1.

\section{Experiments and Results}\label{III_Experiment}
We perform a series of experiments on three publicly available datasets to ascertain the effectiveness of the proposed framework for less-constrained iris recognition. This section firstly provides brief but necessary information for the three public datasets used in this work. We then explain the experimental protocols in the following section. This section also provides a comparative analysis of results from our method with other state-of-the-art methods.
\begin{figure}[!t]
\centering
\includegraphics[width=0.49\textwidth]{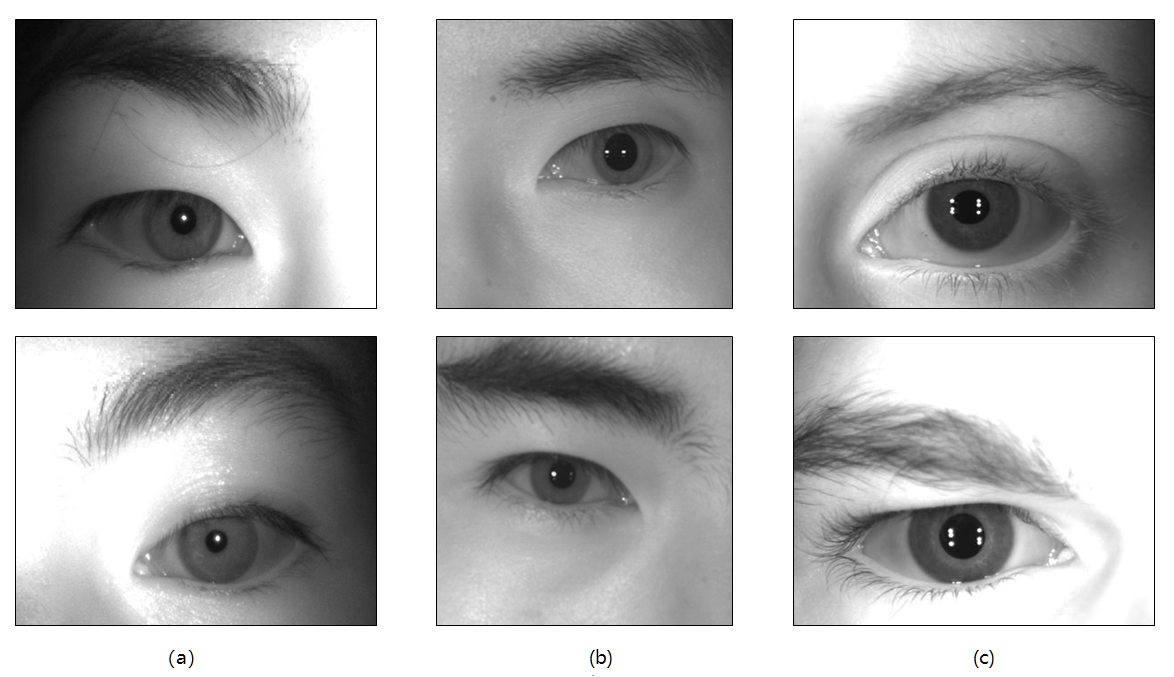}
\caption{ Sample eye images from employed datasets(a) CASIA-Mobile-V1-S3 dataset. (b) CASIA iris image v.4 distance dataset. (c) Q-FIRE-05-middle-illumination dataset.}
\label{fig_sample}
\vspace{-0.5cm}
\end{figure}
\subsection{Datasets and Protocols}
The experimental results presented in this section utilized the following three near-infrared eye image datasets in the public domain. Figure \ref{fig_sample} illustrates the sample eye images from these different datasets.
\subsubsection{Q-FIRE-05-middle-illumination Dataset}
The Quality in Face and Iris Research Ensemble (Q-FIRE) dataset \cite{37JohnsonPeterALopez-MeyerPauloSazonovaNadezhdaHuaFSchuckersS2010} is a publicly available dataset with at-a-distance iris images. Our experiments use Q-FIRE-05-middle-illumination subset which has been acquired at a distance of five feet. under middle-level near-infrared illumination. We automatically segment the periocular region images with a trained Fast-RCNN detector. The processed dataset includes both eye images from 159 different subjects. The first 15 right-eye images are used to train the network while the first ten left-eye images are used for the test evaluation. Therefore this set of experiments generate 7,155 (45 $\times$ 159) genuine scores and 1,256,100 (159 $\times$ 158 $\times$ 50) imposter match scores.  
\subsubsection{CASIA-Moblie-V1-S3 Dataset}
CASIA-Mobile-V1-S3 dataset \cite{34Zhang2018} is another publicly available dataset that includes 3600 face images from 360 different subjects and these images have been acquired using a mobile device with near-infrared illumination. A Fast-RCNN detector \cite{38Girshick2015} is trained with 100 manually labeled samples to detect the periocular region. We follow the same match protocols, both for the iris matching and periocular matching as described in \cite{34Zhang2018}. Therefore the training set includes 3600 samples from 360 classes (eyes) in the first 180 subjects. The test set includes the other 3600 samples from 360 classes (eyes) in 180 subjects. The left eye is matched with all the left-eye images while the right-eye images are matched with all the right ones. After that, the left eye match scores and right eye match scores are combined using the sum rule and generate 8,100 genuine and 1,611,000 imposter match scores.  
\subsubsection{CASIA Iris Image v.4 Distance Dataset}
This subset of the CASIA.v4 database \cite{39BiometricsDatabase} contains the upper part of faces images from 142 subjects. We detect the iris region images with an OpenCV-implemented iris detector \cite{40OpenCVDetector.}, as in earlier references, and generate an eye dataset with 2,446 instances. The training set comprises all the right eye samples, and the test set is composed of all the left eye samples as in \cite{20Zhao2017TowardsFeatures}. The test set therefore generates 20,702 genuine and 2,969,533 imposter match scores.
\vspace{-0.2cm}
\subsection{Iris and Periocular Recognition}\label{III.B within}
\begin{figure}[!t]
\centering
\small
\subfloat[CASIA-Moblie-V1-S3]{\includegraphics[width=2.5in]{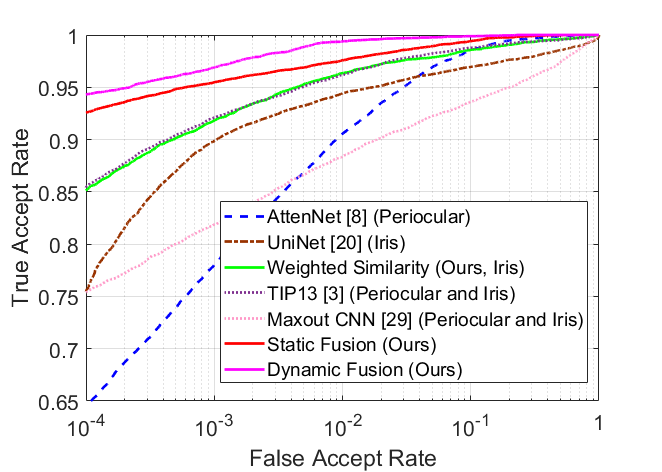}} \\ \vspace{-0.4cm}
\subfloat[ CASIA Iris Image v.4 Distance]{\includegraphics[width=2.5in]{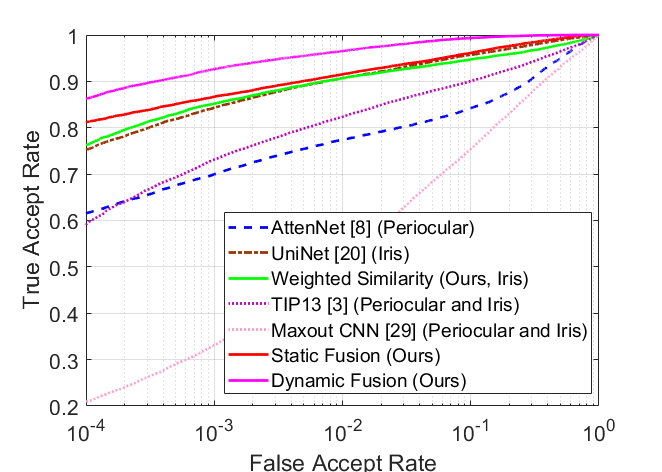}} \\
\vspace{-0.4cm}
\subfloat[Q-FIRE-05-middle-illumination]{\includegraphics[width=2.5in]{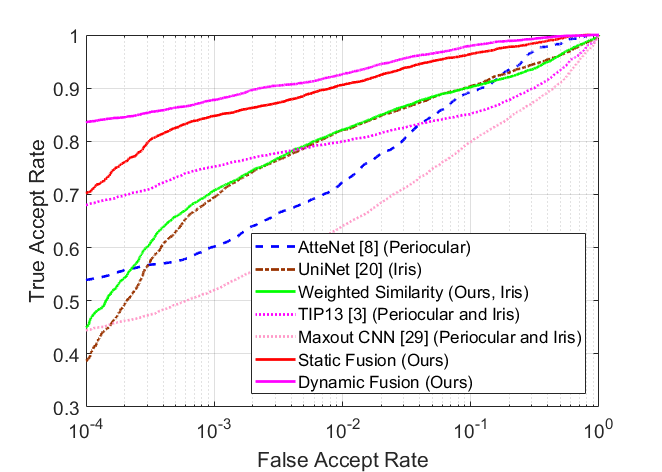}}
\caption{Comparative receiver characteristic curve (ROC) results from within dataset matching.}
\label{fig_within}
\vspace{-0.5cm}
\end{figure}

We firstly present comparative experimental results using simultaneously recovered iris and periocular features using the framework presented in Section \ref{II_method}. Under this set of experiments, all the models were trained using their respective training set and verification performance is evaluated using the respective test set as detailed in earlier sections. We use iris recognition results generated from the \textit{UniNet} \cite{20Zhao2017TowardsFeatures}, and periocular recognition results generated using the \textit{AttenNet} \cite{08Zhao2018}, as the baseline methods for the comparative performance evaluation. Also, we provide comparison using the static score level combination, using the iris match scores generated using similarity measure by us with the periocular match scores, with weighted sum. These comparative results from our algorithms and respective benchmarks are presented in Figure \ref{fig_within} and summarized in Table \ref{table_Within}.
\begin{table*}[!t]
\caption{\normalsize Summary of recognition rates and equal error rates values from the comparison within dataset matching.}
\normalsize
\newcommand{\tabincell}[2]{\begin{tabular}{@{}#1@{}}#2\end{tabular}}
\label{table_Within}
\footnotesize
\centering
\begin{tabular} {|m{3cm}<{\centering}| m{2cm}<{\centering} |m{2cm}<{\centering} |m{2cm}<{\centering} | m{2cm}<{\centering} | m{2 cm}<{\centering} | m{2cm}<{\centering} | }
\hline ~ & \multicolumn{2}{|c|}{CASIA-Mobile-V1-S3} & \multicolumn{2}{|c|}{CASIA.v4 Distance} & \multicolumn{2}{|c|}{Q-FIRE} \\
\hline ~ & TAR$@$FAR=$10^{-4}$ & EER & TAR$@$FAR=$10^{-4}$ & EER  & TAR$@$FAR=$10^{-4}$ & EER \\
\hline 
\tabincell{c}{TIP13\cite{03Tan2013}\\ (Periocular and Iris)} & 85.4\% &2.43\%& 59.2\%& 9.93\%& 68.0\% & 13.86\% \\
\hline
\tabincell{c}{Maxout CNN\cite{34Zhang2018} \\ (Periocular and Iris)} & 75.4\% &7.15\%& 21.0\%& 17.99\%& 44.5\% & 16.74\% \\
\hline
\textit{AttenNet}\cite{08Zhao2018} (Periocular) & 64.6\% & 3.93\%& 61.6\%& 14.27\%& 53.9\% & 10.55\% \\
\hline
\textit{UniNet}\cite{08Zhao2018} (Iris) & 75.5\% & 3.94\%& 75.3\%& 5.54\%& 38.7\% & 9.72\% \\
\hline
\tabincell{c}{Weighted Similarity \\ (Ours, Iris)} & 85.3\% & 2.57\%& 76.0\%& 6.12\%& 44.9\% & 9.85\% \\
\hline
Static Fusion (Ours) & 92.5\% & 1.85\%& 81.5\%& 5.23\%& 69.8\% & 4.95\% \\
\hline
Dynamic Fusion (Ours) & 94.3\% & 0.73\%& 86.3\%& 2.29\%& 83.6\% & 3.97\% \\
\hline
\end{tabular}
\vspace{-0.5cm}
\end{table*}
\par The receiver characteristic curves (ROC) shown in Figure \ref{fig_within}, along with the GAR and equal error rate (EER) summarized in Table \ref{table_Within}, indicate outperforming results in this set of within database experiments. It can be observed that the iris recognition \textit{itself}, using the proposed similarity measure, achieve significantly superior performance over the state of the art iris recognition approach in \cite{20Zhao2017TowardsFeatures}. The combination of respective iris and periocular match scores using static fusion offers significant performance improvement while the dynamic fusion framework using DCNN provides consistently outperforming results on three different datasets. Our approach also outperforms the framework proposed in TIP13 \cite{03Tan2013}. This limited performance can be attributed to the lack of any specialized periocular matching algorithm in \cite{03Tan2013} and our analysis indicates that it is the main constraint in limiting the overall performance. The Maxout CNN is implemented by ourselves based on the parameters provided in \cite{34Zhang2018} since there is no publicly available code for the employed DCNN model and the segmentation algorithm. Also, the Bath dataset used to pre-train the model is no longer publicly available.  
\vspace{-0.2cm}
\subsection{Cross-Database Performance Evaluation} \label{III.C cross}

\begin{table}[!t]
\caption{\normalsize Comparative summary of recognition rates and equal error rates values from the cross-dataset matching.}
\normalsize
\newcommand{\tabincell}[2]{\begin{tabular}{@{}#1@{}}#2\end{tabular}}
\label{table_cross}
\footnotesize
\centering
\begin{tabular} {|m{1.5 cm}<{\centering}| m{1.8cm}<{\centering} |m{1cm}<{\centering} |m{1.5cm}<{\centering} | m{1cm}<{\centering} | }
\hline
Train & \multicolumn{2}{|c|}{CASIA.v4 Distance} & \multicolumn{2}{|c|}{CASIA-Mobile-V1-S3} \\
\hline
Test & CASIA-Mobile-V1-S3 & Q-FIRE & CASIA.v4 Distance & Q-FIRE \\
\hline
\textit{AttenNet} \cite{08Zhao2018} (Periocular) & 9.78\% & 10.15\% & 13.69\% & 12.79\% \\
\hline
\textit{UniNet} \cite{20Zhao2017TowardsFeatures} (Iris) & 4.11\% & 9.72\% & 7.06\% & 10.01\% \\
\hline
Dynamic Fusion (Ours) &1.62\%& 6.49\%& 6.28\% & 6.43\% \\
\hline
 \end{tabular}
\end{table}
\begin{figure*}[!t]
\centering
\small
\subfloat[CASIA-Mobile-V1-S3]{\includegraphics[width=2.5in]{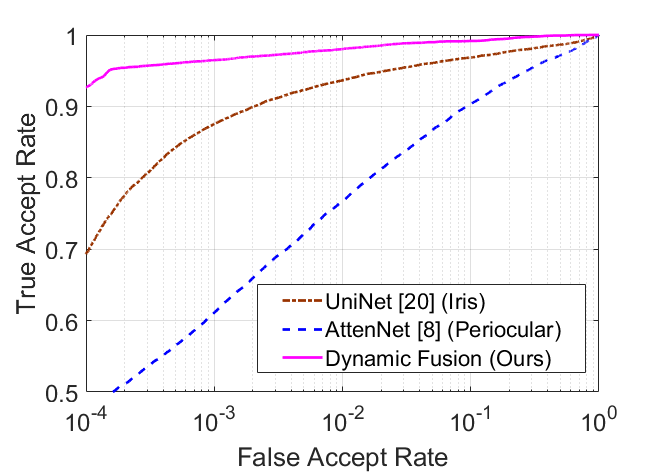} } 
\subfloat[Q-FIRE]{\includegraphics[width=2.5in]{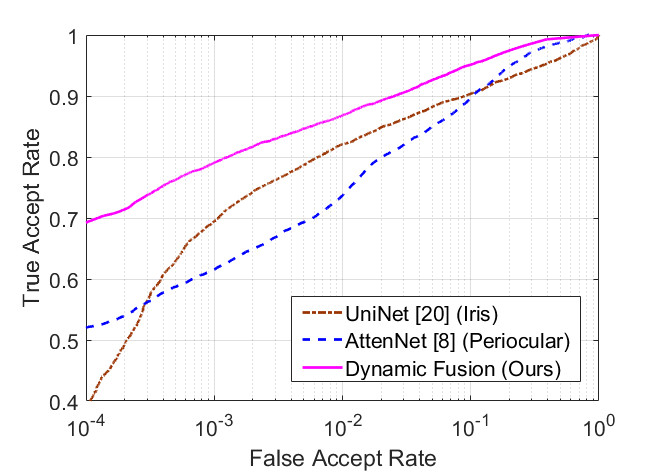}} \\
\vspace{-0.3cm}
\subfloat[CASIA.v4 Distance]{\includegraphics[width=2.5in]{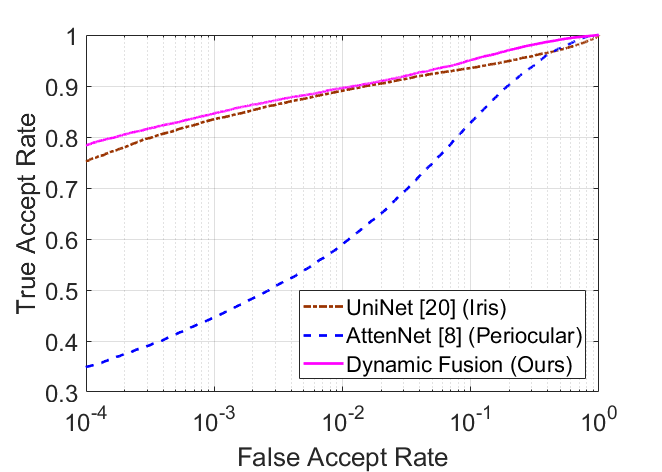}} 
\subfloat[Q-FIRE]{\includegraphics[width=2.5in]{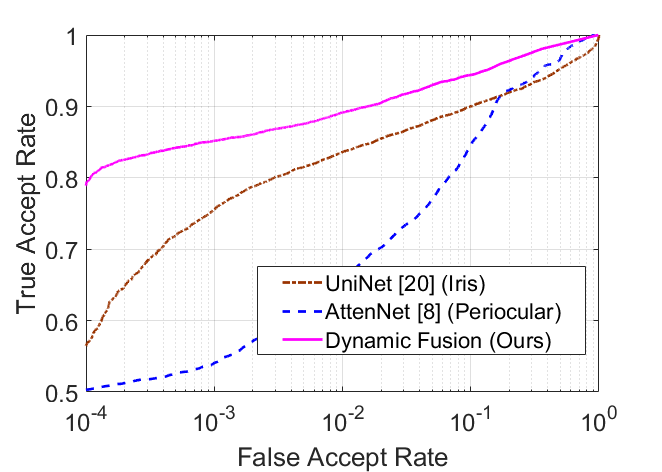}}
\caption{Comparative ROC results from \textit{cross-dataset} performance evaluation. The results in (a)-(b) use the model trained with CASIA.v4 Distance dataset while results in (c)-(d) use the model trained with CASIA-Mobile-V1-S3 dataset.}
\vspace{-0.3cm}
\label{fig_cross}
\end{figure*}

In the cross-database configuration, we incorporate the model trained from CASIA.v4-distance to match CASIA-Mobile-V1-S3 and Q-FIRE dataset images directly \textit{without} any fine-tuning. In addition, we also present cross-database experimental results with the model trained using CASIA-Mobile-V1-S3 database and tested on the CASIA.v4-distance and Q-FIRE dataset images. These set of experiments are aimed to validate the generalization capability of our framework, especially when the image samples available for the training are quite limited. The EER values are summarized in Table \ref{table_cross} and the respective ROCs are shown in Figure \ref{fig_cross}.   
\par The results summarized in this set of experiments indicate consistent improvement from our framework during the cross-database matching which reveals the generality of the framework in matching less-constrained iris images.

\section{Discussion} \label{IV.Discussion}
\begin{figure*}[!ht]
\centering
\small
\subfloat[CASIA-Moblie-V1-S3]{\includegraphics[width=2.3in]{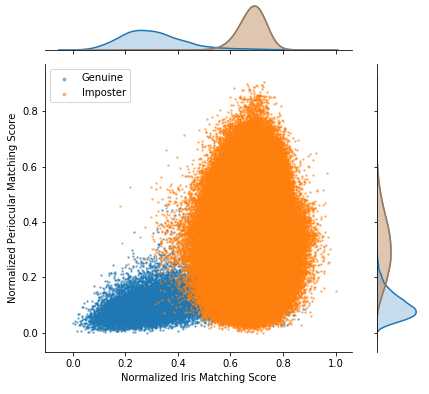}}
\subfloat[CASIA.v4 Distance]{\includegraphics[width=2.3in]{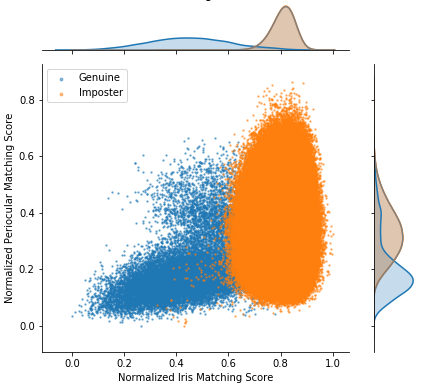}}
\subfloat[Q-FIRE]{\includegraphics[width=2.3in]{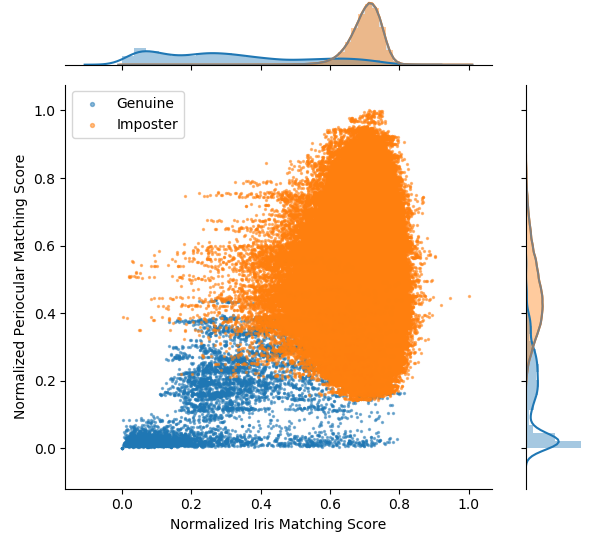}}
\caption{Distribution of matching scores from iris and periocular features.}
\label{fig_dot}
\vspace{-0.8cm}
\end{figure*}

The complementary nature of match scores generated from the deep features in our experiments can be visualized from the two-dimensional plots representing iris and periocular scores. Figure \ref{fig_dot} illustrates such plots for the distribution of (normalized) genuine and imposter scores from iris and periocular matching using respective databases. The subplots in each axis are the kernel density estimation of each score distribution. These plots from less-constrained images indicate that the joint use of individual match scores can be used to more effectively separate genuine and impostor match scores as pursued in this work. 
\begin{figure}[!t]
\centering
\small
\includegraphics[width=3.0in]{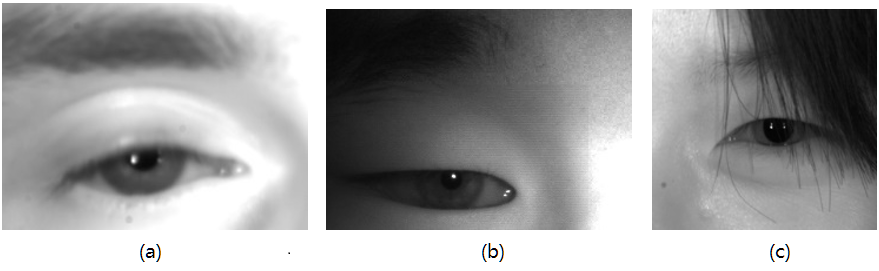}
\caption{Degraded quality image samples: (a) Defocus blur sample in Q-FIRE dataset, (b) Poor illumination sample in CAISA-Mobile-V1-S3 dataset, (c) Severely occluded sample in CASIA v.4 Distance dataset.}
\label{fig_bad_sample}
\vspace{-0.9cm}
\end{figure}

\section{Conclusions and Future Work} \label{V.Conclusion}
This work has introduced a new framework for the periocular assisted iris less-constrained recognition. Our approach has attempted to use better matches for the periocular matching and introduces a similarity score for more accurate iris recognition. The fusion mechanism can dynamically consider the importance of each of the modality, their relative importance, and effective region of the interest to generate more reliable consolidated match scores. The experimental results presented in Section \ref{III_Experiment} using three publicly available datasets demonstrate the merit of the proposed approach, with the outperforming ROC results under within dataset and cross dataset scenario. In order to ensure \textit{reproducibility} of all our results we will provide all codes, along with ground truth labels.  Building an end-to-end framework for periocular and iris recognition framework is one possible direction to further improve this work. Iris recognition itself can be considered as an attention in the periocular recognition. An end-to-end framework that can perform segmentation and simultaneously learn robust features is expected to be more attractive, elegant and is part of further work in this area.

\ifCLASSOPTIONcaptionsoff
  \newpage
\fi



\bibliographystyle{IEEEtran}
%
\bibliography{IEEEexample}


%








\end{document}